%% file: main.tex
\documentclass[10pt,twocolumn,letterpaper]{article}

\usepackage[pagenumbers]{cvpr}

\input{preamble}
\usepackage{algorithm}
\usepackage{algpseudocode}

\definecolor{cvprblue}{rgb}{0.21,0.49,0.74}
\usepackage[pagebackref,breaklinks,colorlinks,allcolors=cvprblue]{hyperref}

\def\confName{CVPR}
\def\confYear{2027}

\title{Latent World Modeling for Point-Cloud Video Pretraining}

\author{Zhuoyue Zhang\textsuperscript{1,2}\quad Jian Liu\textsuperscript{3}\quad Jihua Zhu\textsuperscript{1,2,*}\quad Ajmal Saeed Mian\textsuperscript{4}\\
\textsuperscript{1}Xi'an Jiaotong University\\
\textsuperscript{2}State Key Laboratory of Human-Machine Hybrid Augmented Intelligence\\
\textsuperscript{3}School of Artificial Intelligence and Robotics, Hunan University\\
\textsuperscript{4}University of Western Australia\\
{\small\texttt{zhangzy4016@sru.xjtu.edu.cn}\quad\texttt{zhujh@xjtu.edu.cn}}\\
{\small\textsuperscript{*}Corresponding author}}

\begin{document}
\maketitle
\input{sec/0_abstract}
\input{sec/1_intro}
\input{sec/2_related_work}
\input{sec/3_method}
\input{sec/4_experiments}

\input{sec/5_conclusion}
\input{sec/llm_usage}
{
    \small
    \bibliographystyle{ieeenat_fullname}
    \bibliography{main}
}

\end{document}

%% file: preamble.tex
\newcommand{\std}[1]{{\scriptsize$\pm$#1}}
\newcommand{\diag}[1]{\textcolor{gray}{#1}}
\usepackage{colortbl}
\definecolor{lightblue}{RGB}{221,235,247}
\newcommand{\gain}[1]{{\scriptsize\textcolor{ForestGreen}{$\mathbf{+}$\textbf{#1}}}}
\newcommand{\gainrow}[1]{\quad{\scriptsize\textit{$\Delta$ #1}}}

\newsavebox{\fittabbox}
\newcommand{\fittabular}[3]{%
  \sbox{\fittabbox}{\begin{tabular}{#2}#3\end{tabular}}%
  \setlength{\tabcolsep}{\dimexpr\tabcolsep+(\linewidth-\wd\fittabbox)/#1\relax}%
  \sbox{\fittabbox}{\begin{tabular}{#2}#3\end{tabular}}%
  \setlength{\tabcolsep}{\dimexpr\tabcolsep+(\linewidth-\wd\fittabbox)/#1\relax}%
  \begin{tabular}{#2}#3\end{tabular}}

\makeatletter
\renewcommand{\paragraph}{\@startsection{paragraph}{4}{\z@}%
  {4pt plus 2pt minus 1pt}{-1em}{\normalfont\normalsize\bfseries}}
\makeatother

%% file: sec/0_abstract.tex
\begin{abstract}
Masked modeling is the dominant self-supervised paradigm for point-cloud videos. However, it leaks the information it is meant to predict, whether it reconstructs masked points as in MAE-style methods or predicts their latent features.
Because point-cloud tokens are anchored at points sampled from each frame, querying a target at its own position reveals where its geometry has moved, in reconstruction and latent prediction alike. For latent predictors, random spatio-temporal masks introduce a further shortcut, as visible frames on both sides of each target allow interpolation rather than forecasting.
The pretext task thus reduces to completion. Once the target positions are withheld, a masked latent predictor performs only slightly better than copying the last frame.
To remove this leakage, we propose PointWM, a latent world model that turns completion into forecasting by predicting the features and geometry of future frames from strictly earlier context.
Each target is queried only with its last-seen position and timestamp, while a latent action learned without annotations is meant to capture the part of the future that the past leaves undetermined.
On HOI4D action segmentation with a P4Transformer backbone, PointWM improves F1@50 over training from scratch by 24.9 offline and 29.1 under strictly causal inference.
These results show the promise of latent world models as a leakage-free pretraining paradigm for point-cloud videos.
Our code is publicly available at \url{https://github.com/InitalZ/PointWM}.
\end{abstract}

%% file: sec/1_intro.tex
\section{Introduction}
\label{sec:intro}

Point-cloud videos represent dynamic scenes as sequences of unordered 3D point sets, supporting tasks such as action recognition, temporal action segmentation, and 4D semantic segmentation.
Learning transferable features from these sequences is difficult because annotation is costly and the inputs have neither a regular spatial grid nor persistent point identities.
Although self-supervised pretraining can exploit unlabeled sequences, its value depends on whether the pretext task teaches the representation to model how the scene changes rather than to recover what is already visible.
Predictive objectives are a natural candidate, as illustrated by joint-embedding predictive architectures (JEPAs), which encode visible context and predict the latent features of unseen targets~\cite{assran2023ijepa,bardes2024vjepa,sun2025hyperpoint}. To locate each target, the predictor receives a positional query.

\begin{figure}[t]
  \centering
  \includegraphics[width=\linewidth]{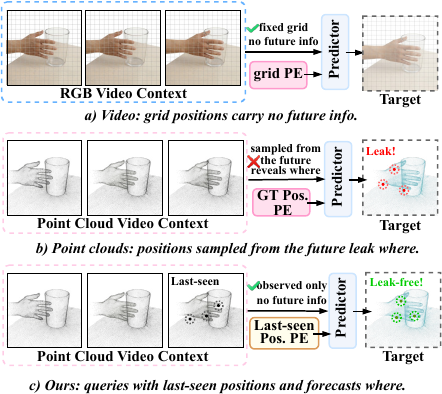}
  \caption{Why target positions matter for point-cloud videos. (a) In RGB video, target positions come from a fixed grid and carry no information about the future. (b) Sampled from the target frame itself, point-cloud anchors have positions that leak where the geometry moves. (c) Using last-seen anchor positions and target timestamps as queries, PointWM forecasts both target features and anchor locations.}
  \label{fig:position_leakage}
\end{figure}

\begin{figure*}[t]
  \centering
  \includegraphics[width=\textwidth]{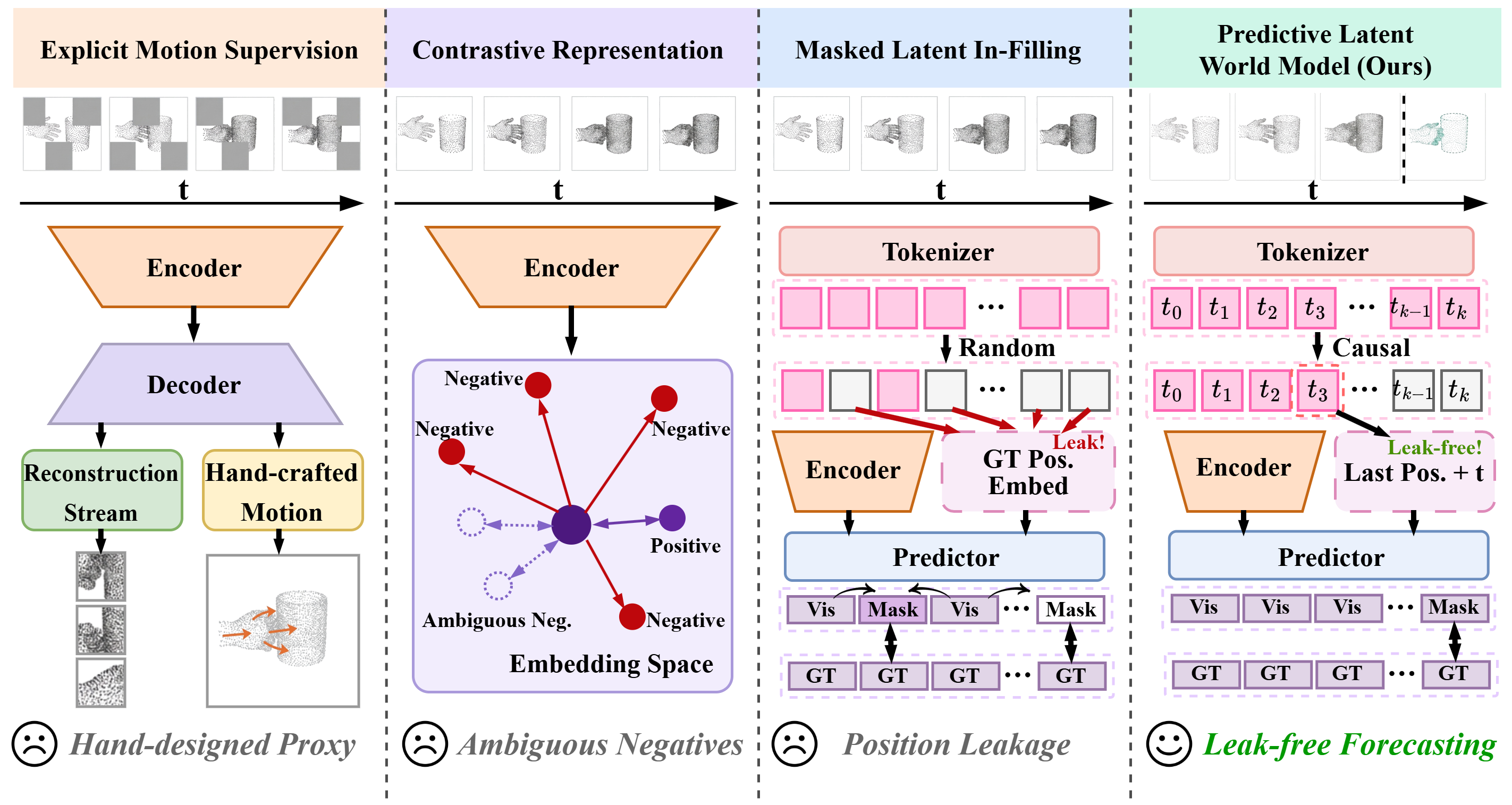}
  \caption{Self-supervised objectives for point-cloud video representation learning, viewed through target-position information. Explicit motion supervision fits hand-designed proxies. Contrastive learning faces ambiguous negatives under temporal redundancy. Masked latent prediction samples targets at random and queries them with their true positions, inheriting the leakage shown in Figure~\ref{fig:position_leakage}(b). With last-seen anchors as queries, PointWM forecasts the features and geometry of future frames from earlier context.}
  \label{fig:motivation_overview}
\end{figure*}

On point-cloud videos, however, this positional query turns prediction into completion.
In RGB video, patch positions lie on a fixed grid, so the query for a future patch is the same whatever the future holds (Figure~\ref{fig:position_leakage}a).
By contrast, point-cloud tokens are anchored at points sampled from the observed surfaces, so a target token's position is computed from the target frame itself and reveals where its geometry has moved (Figure~\ref{fig:position_leakage}b).
We observe this query design in existing latent predictors for point-cloud videos~\cite{lee2026pointtubejepa}.
The resulting trade-off is counterintuitive because a more informative query makes the pretext task easier while weakening the pressure to learn temporal dynamics.
Even withholding target positions is insufficient because random target sampling across the clip leaves visible context on both sides of a target, allowing interpolation rather than forecasting.
A leak-free objective should therefore predict frames that follow the context, with a predictor that knows \emph{when} it is asked to predict but must infer \emph{where} the target geometry will be (Figure~\ref{fig:position_leakage}c).

This perspective also clarifies existing approaches to self-supervised learning on point-cloud videos (Figure~\ref{fig:motivation_overview}).
By recovering masked point tubes together with motion proxies such as cardinality differences, trajectories, or local rotations~\cite{shen2023mastpre,han2024m2psc,wang2026mosaic,sun2026curve3d}, reconstruction-based methods shape the representation through a hand-designed proxy rather than through the motion itself. Because their decoders also receive the true positions of masked tubes, they inherit positional leakage~\cite{shen2023mastpre,han2024m2psc,wang2026mosaic}.
Contrastive methods instead relate clips, frames, or points~\cite{shen2023pointcmp,zuo2026disco4d,li2025pointdico}, or contrast predicted and observed features of later segments~\cite{sheng2023cpr,sheng2023pointcontrast}, but temporal redundancy makes adjacent samples near-duplicates and many negatives ambiguous~\cite{sheng2023pointcontrast}.
Although JEPA-style predictors have reached point-cloud videos~\cite{lee2026pointtubejepa}, they query masked tubes by their true 4D centers and use random masks, inheriting positional leakage and enabling completion rather than forecasting.
To our knowledge, no existing objective for self-supervised learning on point-cloud videos forecasts both the features and the geometry of future frames without access to their positions.

We introduce \textbf{PointWM} (\textbf{P}oint-cloud latent \textbf{W}orld \textbf{M}odeling), a latent world model that forecasts what future frames of a point-cloud video contain and where their geometry moves. To account for what the past leaves undetermined, forecasting is conditioned on a latent action.
PointWM predicts a masked, contiguous window of complete future frames from strictly earlier context, leaving a one-frame gap so that the tokenizer's temporal receptive field does not expose the first target frame.
From queries that combine last-seen anchor positions with target timestamps, a shared predictor outputs both the target features and residual anchor displacements from the last-seen positions.
Because anchors are resampled in every frame, we supervise the geometric output as a set with a symmetric Chamfer objective, without assuming persistent point identities.

\begin{figure*}[t]
  \centering
  \includegraphics[width=\textwidth]{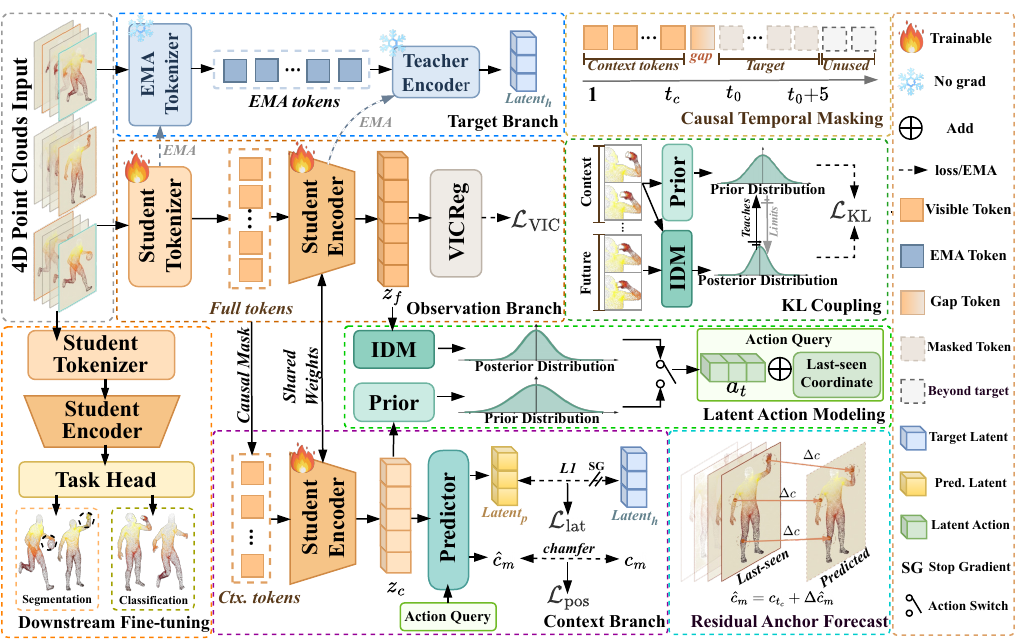}
  \caption{Overview of PointWM.
  \textbf{Target branch.} EMA copies of the tokenizer and encoder encode the full clip without gradients into target latents $\mathbf{h}$ (\emph{Latent}$_h$).
  \textbf{Observation branch.} The student encodes the same full clip into $\mathbf{z}_f$; the latent-action modules read these features, which are regularized by $\mathcal{L}_{\mathrm{VIC}}$.
  \textbf{Context branch.} With causal temporal masking retaining only the frames before a one-frame gap, the weight-shared student encodes these frames into $\mathbf{z}_c$. Queried with last-seen anchor coordinates, target timestamps, and a latent action $\mathbf{a}_t$, the predictor outputs target latents $\hat{\mathbf{h}}$ (\emph{Latent}$_p$) and residual anchor displacements from the last-seen coordinates. The latent predictions are supervised by $\mathcal{L}_{\mathrm{lat}}$ (L1 to the stop-gradient $\mathbf{h}$), and the anchor displacements by the Chamfer loss $\mathcal{L}_{\mathrm{pos}}$.
  \textbf{Latent action modeling.} An inverse dynamics model (IDM) $q_\psi$ infers a posterior over $\mathbf{a}_t$ from the change between pooled frames of $\mathbf{z}_f$. The prior $p_\omega$ is conditioned only on the last context frame of $\mathbf{z}_c$ and the horizon. The loss $\mathcal{L}_{\mathrm{KL}}$ couples them with one-sided stop-gradients, allowing the prior to limit what the posterior encodes while the posterior teaches the prior.
  \textbf{Action switch.} During training, the predictor receives the posterior sample, which is replaced by the prior mean for 30\% of clips. When the future is unobserved, the switch selects the prior.
  Only the student tokenizer and encoder are fine-tuned downstream.}
  \label{fig:pipeline}
\end{figure*}

Withholding target positions makes the task predictive but also ill-posed because the same past can be followed by several futures. A deterministic predictor therefore regresses toward their average.
To address this ambiguity, PointWM conditions forecasting on a compact latent action (Figure~\ref{fig:pipeline}).
In this formulation, a posterior infers the action from the change between frame-pooled student features, while a prior that sees only the context and the forecast horizon learns to imagine it.
The prior limits what the posterior may encode and is in turn trained to match it, which encourages the action to carry what the past does not determine.
Finally, removing the completion objective also removes the implicit regularization it provides. We find that a purely causal objective is prone to collapse, which we counter with variance--covariance regularization on frame-pooled student features.

Our main contributions are as follows.
\begin{itemize}
    \item We identify how existing self-supervised paradigms for point-cloud videos rely on hand-designed proxies, ambiguous negatives, or completion under positional leakage to learn motion without forecasting it. Leak-free forecasting must model how the scene evolves.
    \item We introduce PointWM, the first latent world model for self-supervised learning on point-cloud videos, which forecasts both the features and the anchor geometry of future frames from strictly earlier context, without access to target positions.
    \item To address the ambiguity of forecasting, we introduce an annotation-free latent action whose context-only prior regularizes the posterior toward what the past does not determine.
    \item Experiments on HOI4D action segmentation and MSR-Action3D action recognition show that withholding target positions turns latent prediction into forecasting. PointWM consistently outperforms training from scratch and leaky masked latent prediction, with a 29.1 F1@50 gain over the former in strictly causal online segmentation.
\end{itemize}

%% file: sec/2_related_work.tex
\section{Related Work}
\label{sec:related}

\paragraph{Self-supervised learning on point-cloud videos.}
Reconstruction-based methods extend masked autoencoding to point tubes with hand-designed temporal proxies such as cardinality differences, motion trajectories, or local rotations~\cite{shen2023mastpre,han2024m2psc,wang2026mosaic,han2025rethinking}, but their decoders receive the true positions of masked tubes, leaking target geometry~\cite{sheng2023cpr}.
Whereas contrastive methods discriminate clips, frames, or points~\cite{shen2023pointcmp,zuo2026disco4d,sun2026tri}, contrastive predictive methods go further by using an autoregressor to predict the latent features of a clip's last segment from its earlier segments~\cite{sheng2023cpr,sheng2023pointcontrast}.
The proxies used by reconstruction methods shape the representation instead of the motion itself, while temporal redundancy makes the negatives on which contrastive objectives depend near-duplicate and ambiguous~\cite{sheng2023pointcontrast}. For CPR, embedding the prediction token at the mean position of the target segment also gives its predictions access to the target's centroid~\cite{sheng2023cpr}.
By comparison, PointWM regresses the latent features and anchor geometry of several future frames without proxies or negatives, using only last-seen anchors and target timestamps to query them.

\paragraph{Latent prediction and target-position information.}
Joint-embedding predictive architectures predict target representations from visible context~\cite{assran2023ijepa,bardes2024vjepa} and have been adapted to static point clouds~\cite{saito2025pointjepa,zhang2026cmhanet,liu2026sinref6d,liu2025diff9d} and point-cloud videos. For video, these methods either predict masked point-tube features~\cite{lee2026pointtubejepa} or distill dense targets from frozen image or video foundation models~\cite{nguyen2026cross4djepa}.
Their predictors use position-based queries to locate each target. PointTube-JEPA queries masked tubes by their 4D centers~\cite{lee2026pointtubejepa}, and PointCMP's regressor likewise receives the positions of masked tokens~\cite{shen2023pointcmp}.
Within masked reconstruction of point-cloud videos, DiMP identifies that leaked centers reduce decoding to local geometric retrieval~\cite{zhang2026dimp}. It noises the centers and points of masked tubes and predicts the clean centers from visible context, allowing its decoder to query targets with predicted rather than true positions.
Yet positional queries leak target geometry to latent predictors, and DiMP only noises the leaked information rather than withholding it. At low noise levels, the noised centers and points still reveal the targets almost exactly (Table~\ref{tab:ablation}). Random tube masking across the clip creates another shortcut for latent predictors such as PointTube-JEPA~\cite{lee2026pointtubejepa} because visible context surrounding each target in time allows them to recover masked features through completion rather than forecasting.
PointWM instead withholds target positions entirely, predicting only frames that follow the context and forecasting where their geometry moves. Because its queries use only last-seen anchors and target timestamps, no part of them is computed from the target.

\paragraph{World models and latent actions.}
Developed for pixels and control~\cite{hafner2025dreamerv3,assran2025vjepa2}, latent world models predict future states conditioned on actions.
When action annotations are unavailable, latent actions can be inferred from consecutive observations~\cite{bruce2024genie,schmidt2024lapo,cui2024dynamo,sun2026spikeocc}.
These world models, however, operate on images or videos whose tokens lie on a fixed grid with persistent identities. Both assumptions break in point-cloud videos, where target positions leak future geometry and anchors are resampled in every frame.
To accommodate these differences, PointWM queries with last-seen anchors and supervises geometry as a set without correspondences, while conditioning forecasting on a latent action whose posterior and context-only prior constrain each other.
PointWM is, to our knowledge, the first latent world model for self-supervised learning on point-cloud videos.

%% file: sec/3_method.tex
\section{Method}
\label{sec:method}

PointWM learns a point-cloud-video encoder by forecasting the representations and geometry of future frames from earlier encoded frames, using the three branches of Figure~\ref{fig:pipeline}.

\subsection{Point-Tube Encoding and Causal Masking}
\label{sec:tokenization_masking}
Let $\mathbf{P}=\{\mathbf{P}_t\}_{t=1}^{T}$, where each frame $\mathbf{P}_t\in\mathbb{R}^{N\times 3}$ is an unordered point set.
Following point-cloud-video encoders, PointWM applies farthest-point sampling (FPS) and spatiotemporal ball queries to build local point tubes. A P4DConv tokenizer maps these tubes to tokens $\mathbf{x}_{t,j}\in\mathbb{R}^{d}$ and anchor coordinates $\mathbf{c}_{t,j}\in\mathbb{R}^{3}$, $j\in\{1,\ldots,M\}$, at $T'$ temporal anchors.
The token position is $\mathbf{u}_{t,j}=[\mathbf{c}_{t,j};\tau_t]$, where $\tau_t$ is the normalized temporal index. The variables $\mathbf{x}$ and $\mathbf{u}$ collect all tokens and positions of the clip.
The tokenizer's temporal kernel has size three and stride one, so token $t$ aggregates input from the three frames $t-1$ to $t+1$.

For each clip, we sample a starting anchor $t_0$ and predict $K$ consecutive anchors, $\mathcal{T}=\{t_0,\ldots,t_0+K-1\}$, with all $M$ spatial tokens of each target anchor masked.
The context anchors are $\mathcal{C}=\{t:t<t_0-g\}$ with gap $g=1$. The receptive field of the final context token therefore ends at input frame $t_0-1$, before the first target frame. The intervening anchor belongs to neither set.
We use $K=6$ target anchors and retain at least 12 context anchors.

In the target branch, EMA copies of the tokenizer and encoder, denoted together by $f_{\bar\theta}$, encode the complete clip. The output is layer-normalized ($\operatorname{LN}$) per token.
\begin{equation}
  \mathbf{h}_{t,j}=\operatorname{LN}\!\left(f_{\bar\theta}(\mathbf{P})_{t,j}\right),
  \qquad (t,j)\in\mathcal{T}\times\{1,\ldots,M\}.
  \label{eq:teacher_target}
\end{equation}
The target branch is stop-gradient, with EMA momentum scheduled from 0.996 to 1.0.
In the context branch, the student encoder $f_{\theta}$ sees only tokens indexed by $\mathcal{C}$ and produces the context latents $\mathbf{z}_{c}=f_{\theta}(\mathbf{x}_{\mathcal{C}},\mathbf{u}_{\mathcal{C}})$.

\subsection{Joint Feature and Geometry Forecasting}
\label{sec:joint_prediction}
To avoid leaking target geometry, PointWM forms each target query from the last-seen anchor set and the target time.
Let $t_c=\max\mathcal{C}$ and define $\tilde{\mathbf{c}}_{t,j}=\mathbf{c}_{t_c,j}$ for each target slot $(t,j)$, so the query position is $\tilde{\mathbf{u}}_{t,j}=[\tilde{\mathbf{c}}_{t,j};\tau_t]$.
Because FPS is rerun in every frame, slot $j$ follows the same surface region across nearby anchors only approximately, without exact point correspondence.

The transformer predictor $g_{\phi}$ jointly processes encoded context tokens and target mask tokens at these positions. For each target slot, it outputs a latent prediction $\hat{\mathbf{h}}_{t,j}$ and a residual $\Delta\hat{\mathbf{c}}_{t,j}$.
\begin{equation}
  (\hat{\mathbf{h}}_{t,j},\Delta\hat{\mathbf{c}}_{t,j})
  =g_{\phi}(\mathbf{z}_{c},\mathbf{u}_{\mathcal{C}},
  \tilde{\mathbf{u}}_{t,j},\mathbf{a}_{t}),\qquad
  \hat{\mathbf{c}}_{t,j}=\tilde{\mathbf{c}}_{t,j}+\Delta\hat{\mathbf{c}}_{t,j}.
  \label{eq:predict_features_positions}
\end{equation}
Here $\mathbf{a}_t$ is the latent action of Sec.~\ref{sec:latent_transition}, linearly projected and added to the target query.
The latent objective is the mean absolute error between each slot and the teacher feature of the same-index target anchor, without re-matching,
\begin{equation}
  \mathcal{L}_{\mathrm{lat}}=
  \frac{1}{KM}\sum_{t\in\mathcal{T}}\sum_{j=1}^{M}
  \left\|\hat{\mathbf{h}}_{t,j}-\mathbf{h}_{t,j}\right\|_1.
  \label{eq:feature_loss}
\end{equation}
This approximate alignment suits features, which vary smoothly between nearby anchors, but not coordinates, where slot drift alone incurs large errors. Geometry is therefore supervised with a symmetric set distance that does not depend on slot indices.
\begin{equation}
  \begin{aligned}
    \mathcal{L}_{\mathrm{pos}}=\frac{1}{2K}\sum_{t\in\mathcal{T}}\Bigg[&
      \frac{1}{M}\sum_{j=1}^{M}\min_{j'}
      \|\hat{\mathbf{c}}_{t,j}-\mathbf{c}_{t,j'}\|_2\\
      &+\frac{1}{M}\sum_{j'=1}^{M}\min_{j}
      \|\mathbf{c}_{t,j'}-\hat{\mathbf{c}}_{t,j}\|_2\Bigg].
  \end{aligned}
  \label{eq:position_loss}
\end{equation}

\subsection{Latent Action Posterior and Predictive Prior}
\label{sec:latent_transition}
PointWM represents residual frame-to-frame change with a compact latent action $\mathbf{a}_t$, without action labels.
In the observation branch, the student encodes the full clip as $\mathbf{z}_{f}=f_{\theta}(\mathbf{x},\mathbf{u})$. The posterior state of frame $t$ averages its layer-normalized tokens.
\begin{equation}
  \mathbf{s}_t=\frac{1}{M}\sum_{j=1}^{M}\operatorname{LN}(\mathbf{z}_{f,t,j}),
  \qquad \boldsymbol{\delta}_t=\mathbf{s}_t-\mathbf{s}_{t-1}.
  \label{eq:student_frame_delta}
\end{equation}
An inverse dynamics model (IDM) $q_{\psi}$ maps $\boldsymbol{\delta}_t$ to a diagonal Gaussian posterior $q_t=q_{\psi}(\mathbf{a}_t\mid\boldsymbol{\delta}_t)$ over a $d_a$-dimensional action, with $d_a=8$.
A learned prior $p_t=p_{\omega}(\mathbf{a}_t\mid\mathbf{s}^{c},k)$, also a diagonal Gaussian with mean $\boldsymbol{\mu}^{p}_{t}$, receives the horizon $k=t-t_c$ and the context state $\mathbf{s}^{c}=M^{-1}\sum_{j}\operatorname{LN}(\mathbf{z}_{c,t_c,j})$, pooled from $\mathbf{z}_c$ rather than $\mathbf{z}_f$.
An action switch selects the source of $\mathbf{a}_t$. During training, the posterior sample of a clip is replaced with $\boldsymbol{\mu}^{p}_{t}$ with probability 0.3, exposing the predictor to the action available from context alone. At prediction time, the prior mean is always used because the future is unobserved.

We route the KL gradients of the posterior and the prior separately~\cite{hafner2025dreamerv3}.
\begin{align}
  \mathcal{L}_{q}&=\sum_{i=1}^{d_a}\max\!\left(\mathbb{E}[D_{\mathrm{KL}}(q_{t,i}\|\operatorname{sg}(p_{t,i}))],\lambda_{\mathrm{fb}}\right),\\
  \mathcal{L}_{p}&=\mathbb{E}[D_{\mathrm{KL}}(\operatorname{sg}(q_t)\|p_t)].
  \label{eq:action_kl}
\end{align}
Here $q_{t,i}$ and $p_{t,i}$ are the $i$-th marginals, and $\operatorname{sg}$ denotes stop-gradient. The expectation is over target transitions in a minibatch, and free bits $\lambda_{\mathrm{fb}}$ apply per posterior dimension.
The KL coupling $\mathcal{L}_{\mathrm{KL}}=\beta_q\mathcal{L}_{q}+\beta_p\mathcal{L}_{p}$ lets the prior limit the posterior ($\mathcal{L}_{q}$) and the posterior teach the prior ($\mathcal{L}_{p}$).

\subsection{Training Objective}
\label{sec:objective}
To discourage collapse, we apply the variance and covariance terms of VICReg~\cite{bardes2022vicreg,wang2026pointrft} to the student's frame-pooled features $\bar{\mathbf{z}}_t=M^{-1}\sum_j\mathbf{z}_{f,t,j}$. This matches the pooling of the downstream segmentation head.
Let $\sigma_r$ denote the standard deviation of channel $r$. Let $\mathbf{C}$ denote the channel covariance over pooled features across clips and times.
\begin{equation}
  \mathcal{L}_{\mathrm{VIC}}=\frac{1}{d}\sum_{r=1}^{d}\max(0,1-\sigma_r)
  +\frac{0.04}{d}\sum_{r\ne s}C_{rs}^{2}.
  \label{eq:vicreg_loss}
\end{equation}
The complete pretraining objective is
\begin{equation}
  \mathcal{L}=\mathcal{L}_{\mathrm{lat}}
  +\lambda_{\mathrm{pos}}\mathcal{L}_{\mathrm{pos}}
  +\mathcal{L}_{\mathrm{KL}}
  +\lambda_{\mathrm{VIC}}\mathcal{L}_{\mathrm{VIC}},
  \label{eq:total_loss}
\end{equation}
with $\lambda_{\mathrm{pos}}=5$, $\beta_q=0.05$, $\lambda_{\mathrm{fb}}=0.2$, $\beta_p=0.5$, and $\lambda_{\mathrm{VIC}}=0.1$.
The tokenizer uses 64 FPS anchors per frame (2048 input points). The student encoder has dimension 384, depth 6, and 6 heads. The predictor has dimension 256, depth 4, and 4 heads.
Downstream tasks keep only the student tokenizer and encoder.
Pretraining clips are normalized with full-clip statistics before tokenization. Sec.~\ref{sec:exp_ablation} tests whether forecasting depends on this normalization.
Training settings are given in Sec.~\ref{sec:exp_setup_main}.

%% file: sec/4_experiments.tex
\section{Experiments}
\label{sec:exp}

We first test the central claim that withholding target positions turns latent prediction into forecasting (Sec.~\ref{sec:exp_forecast}). We then ask whether the resulting encoder transfers offline, across datasets (Sec.~\ref{sec:exp_transfer}), and under strictly causal inference (Sec.~\ref{sec:exp_online}).
Finally, we attribute these results to individual components (Sec.~\ref{sec:exp_ablation}).

\subsection{Experimental Setup}
\label{sec:exp_setup_main}

\paragraph{Datasets.}
HOI4D~\cite{liu2022hoi4d} has 2{,}971 training and 892 test egocentric hand--object videos of 150 frames, labeled per frame with 19 actions.
We hold out 10\% of the training videos for model selection and pretrain on the rest without labels, so no test video is seen in pretraining.
MSR-Action3D~\cite{li2010msraction3d} has 567 videos of 20 full-body actions, split by subject into 270 training and 297 test videos.

\paragraph{Implementation.}
We pretrain on 32-frame clips for 100 epochs with AdamW~\cite{loshchilov2019adamw}, batch size 64, and weight decay 0.05. The learning rate warms up over 15 epochs to $5\times10^{-4}$ and cosine-decays to $10^{-6}$. Fine-tuning runs end to end for 30 epochs. The backbone is P4Transformer~\cite{fan2021p4transformer} unless noted, and all models are trained on four A800 GPUs.

\paragraph{Baselines.}
\emph{Scratch} is the same backbone without pretraining. Controlled variants differ from PointWM only in the pretext objective. \emph{Block} trains with block masks and \emph{Leak} adds causal masks, both querying targets at their true positions. \emph{Last} keeps both mask types but queries every target with the anchors of its nearest preceding visible frame, removing only the leakage.

\subsection{Forecasting Without Target Positions}
\label{sec:exp_forecast}

\begin{table}[t]
  \centering
  \footnotesize
  \setlength{\tabcolsep}{3pt}
  \fittabular{16}{lccccccc}{%
    \toprule
    & \multicolumn{4}{c}{Skill vs.\ baseline} & \multicolumn{3}{c}{Feature change} \\
    \cmidrule(lr){2-5}\cmidrule(lr){6-8}
    Pretraining & \diag{CL$^{\ast}$} & CL & CM & LE & $\Delta R^2$ & mov. & cos \\
    \midrule
    Block         & \diag{0.81} & 0.03 & $-$0.35 & $-$0.12 & 0.05 & 0.04 & 0.26 \\
    Leak          & \diag{0.85} & $-$0.07 & $-$0.18 & $-$0.12 & $-$0.08 & $-$0.12 & 0.09 \\
    Teacher act.  & --          & 0.27 & 0.15    & 0.12    & 0.30 & 0.26 & 0.52 \\
    Last          & --          & 0.42 & 0.29 & 0.25 & 0.47 & 0.39 & 0.75 \\
    \rowcolor{lightblue}
    PointWM      & \diag{0.54} & \textbf{0.52} & \textbf{0.40} & \textbf{0.37} & \textbf{0.58} & \textbf{0.49} & \textbf{0.81} \\
    \bottomrule
  }
  \caption{Latent forecasting by frozen pretrained models on HOI4D, averaged over $t{+}1$ to $t{+}6$.
  Skill against copy-last (CL), the context mean (CM), and linear extrapolation (LE) is $(\cos_{\text{model}}-\cos_{\text{base}})/(1-\cos_{\text{base}})$.
  \diag{CL$^{\ast}$} queries with the true target anchors as a leakage diagnostic.
  $\Delta R^2$ is the explained variance of the feature change over all and moving (mov.) tokens, and cos is the direction cosine between predicted and true change.
  Teacher act. reads the latent action from EMA-teacher frame differences.}
  \label{tab:forecast}
\end{table}

\begin{figure}[t]
  \centering
  \includegraphics[width=\linewidth]{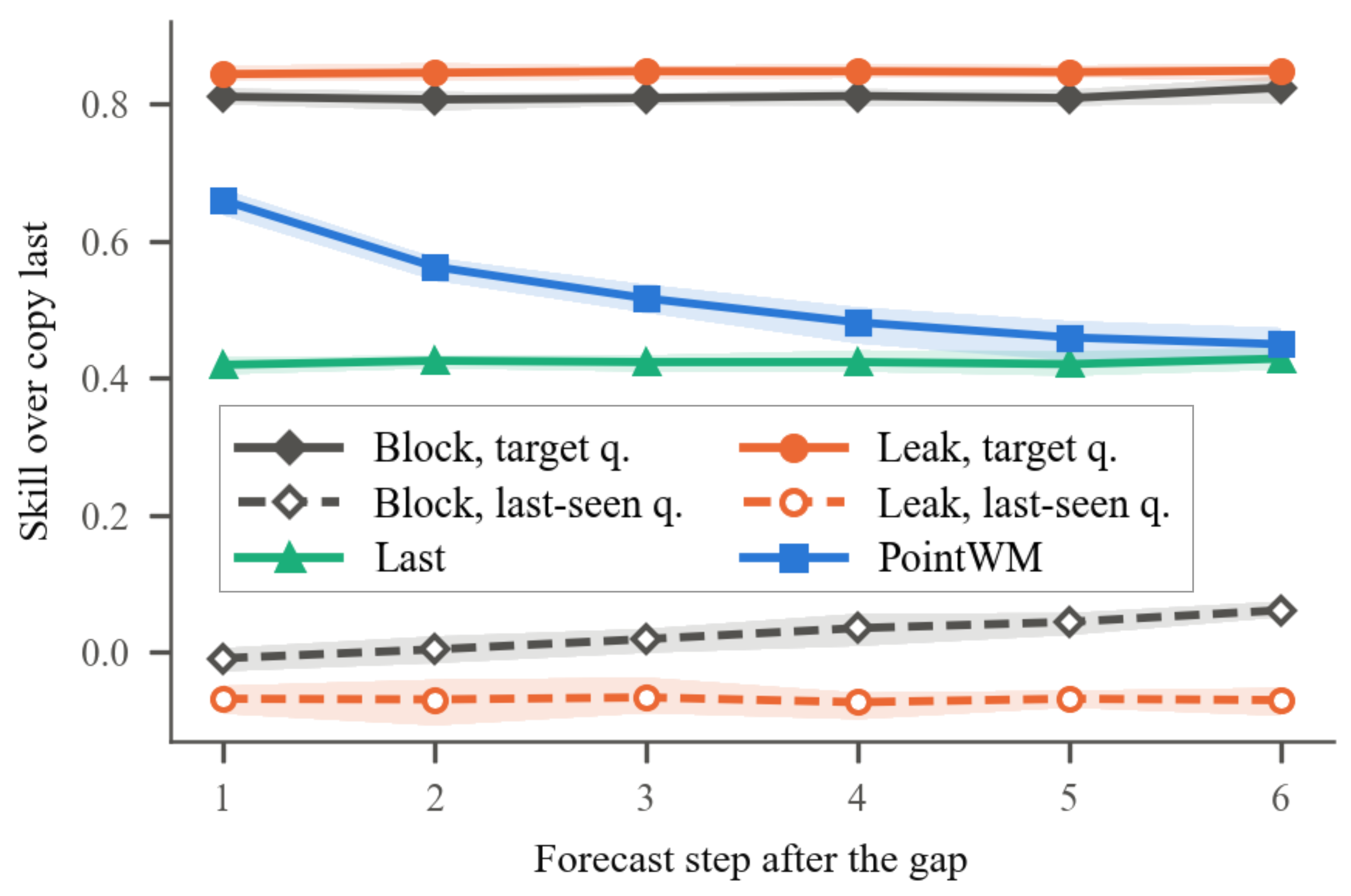}
  \caption{Forecast skill over copy-last versus forecast step on HOI4D. Shading denotes the 95\% CI over clips, and 0 corresponds to copying the last context frame.
  Each color denotes one model. Block and Leak are queried at test time with target positions (solid) or last-seen anchors (dashed), while Last and PointWM are queried with last-seen anchors.
  Block uses block masks only, and Leak adds causal masks. Both are trained with target-position queries.}
  \label{fig:horizon}
\end{figure}

We probe each frozen pretrained model on the HOI4D test split.
The predictor receives the first 25 anchor frames as context and forecasts the six frames after a one-frame gap, queried with last-seen anchors and conditioned on the prior-mean action.
Because adjacent frames are nearly identical, we report skill relative to three predictors that require no dynamics: copy-last, the context mean, and linear extrapolation.
We also measure how much of the actual feature change $\mathbf{h}_{t}-\mathbf{h}_{t_c}$ is explained ($\Delta R^2$). A predictor that only reproduces the static scene scores zero on it.
Targets and baselines all use each model's own full-clip teacher features, paired by slot, so these metrics measure forecasting within that space rather than physical motion.

\paragraph{Leaked positions do the predicting.}
Queried with target positions, both completion models reach high skill over copy-last (Table~\ref{tab:forecast}). Block reaches 0.81 despite never being trained to forecast, and Leak reaches 0.85.
Given last-seen anchors instead, Block drops to 0.03, below the context mean ($-0.35$) and linear extrapolation ($-0.12$). Leak drops to $-0.07$, worse than copying the last frame. Neither explains more than 5\% of the feature change.
Their apparent skill is read off the target geometry in the query.
Figure~\ref{fig:horizon} shows the same signature across horizons. With target positions, skill is flat from $t{+}1$ to $t{+}6$, as expected when the answer is supplied rather than inferred.
Adding a causal objective while the leak remains only deepens this dependence.
PointWM, in contrast, gains only 0.02 when given the target positions (0.54 vs.\ 0.52), against 0.78 for Block and 0.92 for Leak, so its forecast does not rely on them.

\begin{table}[t]
  \centering
  \footnotesize
  \setlength{\tabcolsep}{3pt}
  \fittabular{12}{lccccc}{%
    \toprule
    Method & Acc & Edit & F1@10 & F1@25 & F1@50 \\
    \midrule
    PPTr~\cite{wen2022pptr}          & 77.4 & 80.1 & 81.7 & 78.5 & 69.5 \\
    \;+ STRL~\cite{huang2021strl}      & 78.4 & 79.1 & 81.8 & 78.6 & 69.7 \\
    \;+ VideoMAE~\cite{tong2022videomae}  & 78.6 & 80.2 & 81.9 & 78.7 & 69.9 \\
    \;+ C2P~\cite{zhang2023c2p}       & 81.1 & 84.0 & 85.4 & 82.5 & 74.1 \\
    \;+ DiMP~\cite{zhang2026dimp}      & 84.0 & 86.2 & 89.1 & 86.6 & 81.4 \\
    \rowcolor{lightblue}
    \;+ PointWM   & \textbf{84.2}\std{0.3} & \textbf{88.0}\std{0.8} & \textbf{92.3}\std{0.5} & \textbf{89.4}\std{0.4} & \textbf{84.0}\std{0.5} \\
    \gainrow{vs.\ Scratch} & \gain{6.8} & \gain{7.9} & \gain{10.6} & \gain{10.9} & \gain{14.5} \\
    \midrule
    P4Transformer~\cite{fan2021p4transformer} & 71.2 & 73.1 & 73.8 & 69.2 & 58.2 \\
    \;+ C2P~\cite{zhang2023c2p}       & 73.5 & 76.8 & 77.2 & 72.9 & 62.4 \\
    \;+ MaST-Pre~\cite{shen2023mastpre}$^\dagger$ & 74.1 & 75.4 & 76.6 & 73.4 & 63.8 \\
    \;+ M2PSC~\cite{han2024m2psc}$^\dagger$    & 75.9 & 77.1 & 78.4 & 74.1 & 65.9 \\
    \;+ DiMP~\cite{zhang2026dimp}      & 82.4 & 84.9 & 87.3 & 85.9 & 80.8 \\
    \rowcolor{lightblue}
    \;+ PointWM   & \textbf{82.5}\std{0.4} & \textbf{87.2}\std{0.5} & \textbf{90.4}\std{0.4} & \textbf{88.7}\std{0.4} & \textbf{83.1}\std{0.6} \\
    \gainrow{vs.\ Scratch} & \gain{11.3} & \gain{14.1} & \gain{16.6} & \gain{19.5} & \gain{24.9} \\
    \bottomrule
  }
  \caption{Offline action segmentation on HOI4D (150-frame inputs).
  Rows marked with $\dagger$ are unofficial results.
  PointWM (shaded) reports the mean\,$\pm$\,std over three seeds. Each seed is selected on validation F1@50 and evaluated once on the 892 test videos.
  Gains compare PointWM with the backbone without pretraining (the first row of each block).}
  \label{tab:hoi4d_seg}
\end{table}

\begin{figure}[t]
  \centering
  \includegraphics[width=\linewidth]{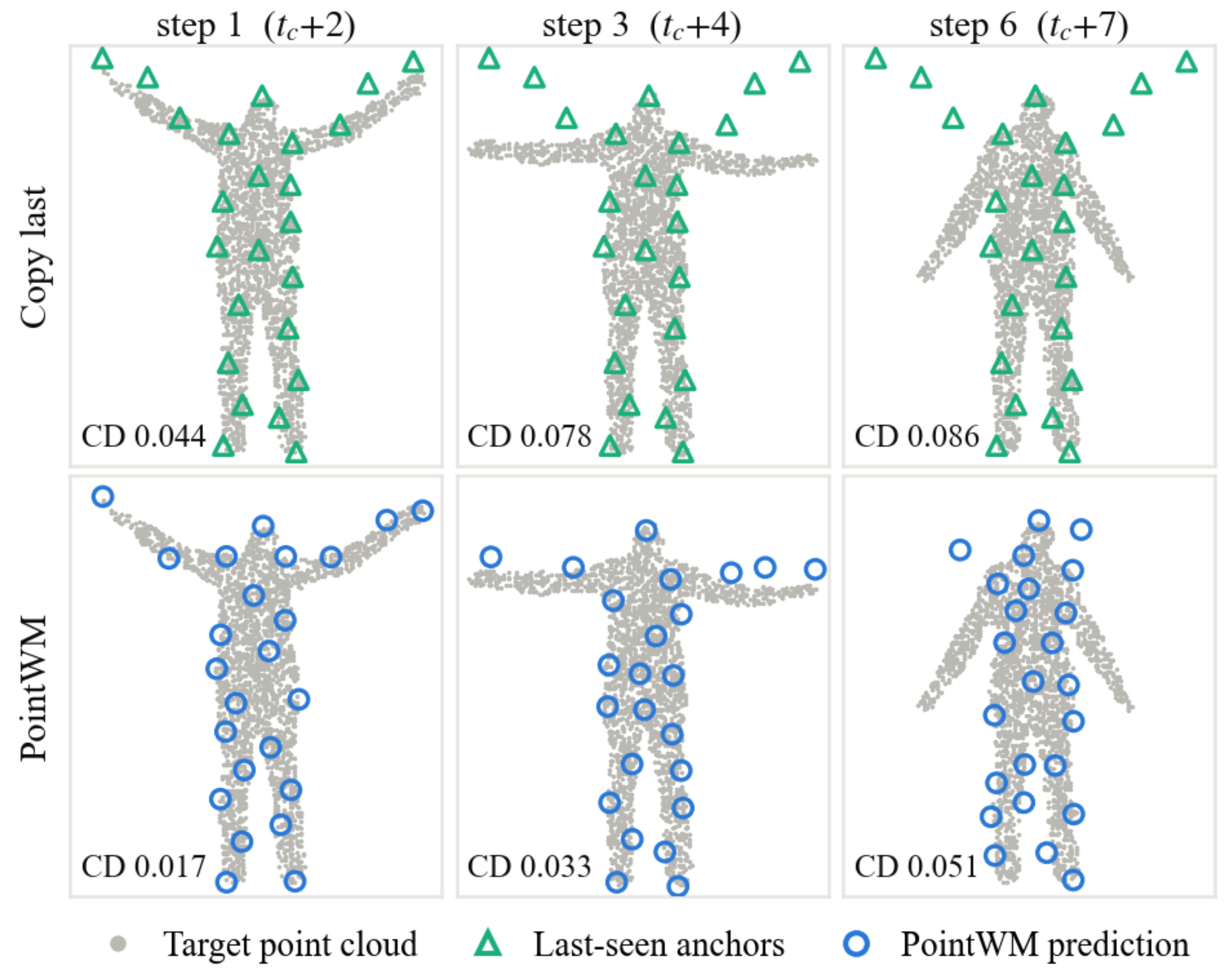}
  \caption{Anchor forecasts at three forecast steps.
  Gray points show the raw target point cloud.
  The top row shows last-seen anchors (copy-last), and the bottom row shows PointWM's position-head forecast.}
  \label{fig:anchors}
\end{figure}

\paragraph{PointWM predicts feature change, not only the static scene.}
PointWM beats copy-last at every horizon and explains 58\% of the change in its target features. It explains 49\% on moving tokens, with a direction cosine of 0.81.
Last improves on Leak but trails PointWM by 0.10 in skill and 0.11 in $\Delta R^2$. Removing the leakage is necessary but not sufficient.
The two also differ in shape (Figure~\ref{fig:horizon}). PointWM's skill decays from 0.66 at $t{+}1$ to 0.45 at $t{+}6$, as expected when the forecast must extrapolate further. Last stays near 0.42 at every step.
Reading the latent action from the EMA teacher lowers skill below Last (0.27). The action helps only when its gradient reaches the student encoder.
In $\Delta R^2$, PointWM decays from 0.69 at $t{+}1$ to 0.50 at $t{+}6$. Rolling the predictor out autoregressively decays faster, reaching 0.27 at $t{+}6$. We therefore forecast all horizons in parallel.
Figure~\ref{fig:anchors} shows position-head anchor forecasts.

\subsection{Downstream Transfer}
\label{sec:exp_transfer}

\begin{table}[t]
  \centering
  \footnotesize
  \setlength{\tabcolsep}{2pt}
  \begin{tabular}{@{}lc|lc@{}}
    \toprule
    Method & Acc & Method & Acc \\
    \midrule
    PSTNet~\cite{fan2021pstnet}           & 91.20 & P4Transformer~\cite{fan2021p4transformer} & 90.94 \\
    \;+ PointCMP~\cite{shen2023pointcmp}     & 93.27 & \;+ MaST-Pre~\cite{shen2023mastpre} & 91.29 \\
    PST-Transformer~\cite{fan2022psttransformer}  & 93.73 & \;+ M2PSC~\cite{han2024m2psc}     & 93.03 \\
    \;+ MaST-Pre~\cite{shen2023mastpre}     & 94.08 & \;+ Disco4D~\cite{zuo2026disco4d} & 93.38 \\
    \;+ M2PSC~\cite{han2024m2psc}        & 94.84 & \;+ DiMP~\cite{zhang2026dimp}      & 93.97 \\
    \;+ DiMP~\cite{zhang2026dimp}         & 95.51 & \;+ PointTube-JEPA~\cite{lee2026pointtubejepa} & 94.08 \\
    \rowcolor{lightblue}
    \;+ PointWM (HOI4D) & \textbf{96.18}\std{0.19} & \;+ PointWM (HOI4D) & 95.06\std{0.39} \\
    \gainrow{vs.\ Scratch} & \gain{2.45} & \gainrow{vs.\ Scratch} & \gain{4.12} \\
    \bottomrule
  \end{tabular}
  \caption{Action recognition accuracy (\%) on MSR-Action3D.
  PointWM (shaded) is pretrained on HOI4D, fine-tuned end to end, and evaluated at the last epoch on all 297 test videos. Results report the mean\,$\pm$\,std over three seeds.
  Gains compare PointWM with the backbone without pretraining (the PST-Transformer and P4Transformer rows).}
  \label{tab:msr}
\end{table}

\paragraph{HOI4D action segmentation.}
On the P4Transformer backbone, PointWM raises F1@50 from 58.2 to 83.1 (+24.9) and Edit by 14.1 over training without pretraining (Table~\ref{tab:hoi4d_seg}). On PPTr, it raises F1@50 by 14.5 and Edit by 7.9.
Within our framework, PointWM exceeds Leak by 15.0 F1@50 when only 10\% of the training videos are labeled (Table~\ref{tab:ablation}, Seg$_{10\%}$), so withholding target positions matters for the offline task as well.

\paragraph{Transfer to MSR-Action3D.}
Fine-tuned from the HOI4D checkpoint, PointWM improves accuracy on MSR-Action3D by 4.12 on P4Transformer (90.94\% to 95.06\%) and by 2.45 on PST-Transformer (93.73\% to 96.18\%) over training without pretraining (Table~\ref{tab:msr}).
Within our framework, it exceeds Leak by 2.69 on P4Transformer (Table~\ref{tab:ablation}, MSR).
This suggests that forecasting hand--object interactions yields features that carry over to full-body actions recorded in a different dataset.

\subsection{Causal Online Segmentation}
\label{sec:exp_online}

\begin{table}[t]
  \centering
  \footnotesize
  \setlength{\tabcolsep}{3pt}
  \begin{tabular}{@{}lccccc@{}}
    \toprule
    Method & Acc & Edit & F1@10 & F1@25 & F1@50 \\
    \midrule
    PPTr~\cite{wen2022pptr}          & 69.7 & 64.5 & 69.4 & 64.8 & 52.9 \\
    P4Mamba~\cite{liu2026pointnet4d}       & 71.0 & 78.5 & 77.8 & 73.4 & 61.5 \\
    PointNet4D~\cite{liu2026pointnet4d}    & 70.3 & 72.3 & 73.7 & 69.1 & 57.2 \\
    4DMAP~\cite{liu2026pointnet4d}         & 72.4 & 78.7 & 78.8 & 74.8 & 63.6 \\
    P4Transformer~\cite{fan2021p4transformer} & 66.7 & 62.0 & 65.3 & 59.8 & 46.3 \\
    \;+ DiMP~\cite{zhang2026dimp}      & 80.4 & 82.2 & 84.8 & 83.2 & 73.5 \\
    \;+ Leak     & 70.2\std{0.7} & 64.8\std{1.1} & 73.5\std{0.8} & 67.3\std{1.3} & 56.1\std{0.9} \\
    \;+ Last      & 76.8\std{0.9} & 76.9\std{1.2} & 81.9\std{1.1} & 77.7\std{0.9} & 68.6\std{0.8} \\
    \rowcolor{lightblue}
    \;+ PointWM   & \textbf{81.0}\std{0.4} & \textbf{83.0}\std{0.6} & \textbf{85.9}\std{0.7} & \textbf{84.4}\std{1.1} & \textbf{75.4}\std{0.8} \\
    \gainrow{vs.\ Scratch} & \gain{14.3} & \gain{21.0} & \gain{20.6} & \gain{24.6} & \gain{29.1} \\
    \gainrow{vs.\ Leak}    & \gain{10.8} & \gain{18.2} & \gain{12.4} & \gain{17.1} & \gain{19.3} \\
    \bottomrule
  \end{tabular}
  \caption{Strictly causal online action segmentation on HOI4D (150-frame sequences, no look-ahead).
  Our runs (Leak, Last, PointWM) report the mean\,$\pm$\,std over three seeds. PointWM is shaded.
  Gains compare PointWM with the backbone without pretraining (the P4Transformer row) and with Leak.}
  \label{tab:online}
\end{table}

Following the causal protocol of PointNet4D~\cite{liu2026pointnet4d,sun2026align}, we process each 150-frame sequence in its raw HOI4D coordinates with a causal attention mask and a causal P4DConv temporal kernel, so token $t$ aggregates frames $t{-}2$ to $t$. The label of frame $t$ depends only on frames up to $t$, with no smoothing.
PointWM improves on Scratch by 14.3 accuracy and 29.1 F1@50 and on Leak by 10.8 accuracy and 19.3 F1@50 (Table~\ref{tab:online}).
The controlled variants are ordered as predicted by the forecasting probe in Table~\ref{tab:forecast}.
Leak already improves on Scratch by 9.8 F1@50. Removing the leakage (Last) adds a further 12.5, the largest single step, mirroring the largest step in forecast skill ($-0.07$ to 0.42).
The full objective adds another 6.8 over Last, which Sec.~\ref{sec:exp_ablation} attributes to individual components.
Under the causal constraint, PointWM loses 7.7 F1@50 and Scratch loses 11.9, so the gain over Scratch grows from 24.9 offline (Table~\ref{tab:hoi4d_seg}) to 29.1 online.
Forecasting pretraining thus helps most when the model sees only the past.

\subsection{Ablation Studies}
\label{sec:exp_ablation}

\begin{table}[t]
  \centering
  \footnotesize
  \setlength{\tabcolsep}{3pt}
  \fittabular{14}{llccccc}{%
    \toprule
    & & \multicolumn{2}{c}{Pretext} & \multicolumn{3}{c}{Downstream} \\
    \cmidrule(lr){3-4}\cmidrule(l){5-7}
    & Pretraining & Skill & Rank & Seg$_{10\%}$ & Online & MSR \\
    \midrule
    (a) & Leak                     & $-$0.07 & 94.3 & 54.2 & 56.1 & 92.37 \\
    (b) & Last                     & 0.42 & 138.7 & 63.4 & 68.6 & 93.94 \\
    (c) & (b) $-$ block masks      & 0.04 & 12.6 & 48.6 & 52.8 & 87.09 \\
    (d) & (c) $+$ frame VICReg     & 0.43 & 178.2 & 65.1 & 69.4 & 93.60 \\
    (e) & (d) $+$ position head    & 0.48 & 182.6 & 67.0 & 72.5 & 94.39 \\
    \rowcolor{lightblue}
    (f) & (e) $+$ latent action    & \textbf{0.52} & 185.1 & \textbf{69.2} & \textbf{75.4} & \textbf{95.06} \\
    \midrule
    (g) & (f), gap $g{=}0$         & 0.46 & 176.8 & 67.7 & 73.9 & 94.73 \\
    (h) & (f), token VICReg        & 0.44 & 226.4 & 64.3 & 70.1 & 93.94 \\
    \midrule
    (i) & (f), target query        & 0.16 & 181.7 & 62.9 & 67.8 & 93.60 \\
    (j) & (f), noised, small $\sigma$ & 0.24 & 189.3 & 64.5 & 69.6 & 93.83 \\
    (k) & (f), noised, large $\sigma$ & 0.44 & 180.8 & 67.2 & 73.0 & 94.50 \\
    (l) & (f), time-only query     & 0.36 & 186.4 & 65.8 & 71.1 & 94.16 \\
    \bottomrule
  }
  \caption{Component ablation. Each row changes one factor of the row it names. Row (f) is PointWM, and rows (g) to (l) are variants of it.
  Rows (i) to (l) change only the target query ($\sigma=0.02$ / $0.20$).
  Skill denotes forecast skill over copy-last, averaged over $t{+}1$ to $t{+}6$ with the protocol of Table~\ref{tab:forecast}.
  Rank denotes the effective rank of frame-pooled student features on the HOI4D test split.
  Seg$_{10\%}$ denotes offline HOI4D action segmentation F1@50 with 10\% of the training videos labeled.
  Online denotes causal F1@50 with the protocol of Table~\ref{tab:online}.
  MSR denotes MSR-Action3D accuracy.}
  \label{tab:ablation}
\end{table}

Table~\ref{tab:ablation} adds the components of PointWM one at a time, while Table~\ref{tab:latent_action} varies the design of the latent action.

\paragraph{Components.}
Removing the leakage (a$\to$b) gives the largest gain, raising skill from $-0.07$ to 0.42 and online F1@50 by 12.5.
A purely causal objective (c) then collapses. The effective rank falls from 138.7 to 12.6, and every downstream metric drops below Leak. Frame-level VICReg (d) prevents the collapse and recovers the performance of Last without completion.
The position head (e) and the latent action (f) each add about 3 online F1@50, together accounting for 6.0 of the 6.8 by which PointWM exceeds Last.
Removing the one-frame gap (g) or applying VICReg per token (h) lowers every downstream metric. The latter lowers online F1@50 by 5.3 despite a higher rank.

\paragraph{Query design.}
Rows (i) to (l) keep every component of PointWM and change only the target query. With true target anchors (i), PointWM falls below Last on every downstream metric, so withholding the target positions outweighs all other components combined. Gaussian noise removes the leak only in part, reaching 69.6 (j) and 73.0 (k) online F1@50 against 75.4. A time-only query (l) reaches 71.1, above the leaky queries but below last-seen anchors, which thus also act as a spatial scaffold. Skill and all downstream metrics follow the same order across these rows, unlike the effective rank.

\paragraph{Latent action.}
A fixed $\mathcal{N}(\mathbf{0},\mathbf{I})$ prior leaves the prior forecast no better than having no action (0.40). An IDM that also reads the previous state has a between-clip variance of 71.4\%, suggesting that it encodes scene identity, and does not help segmentation.
Only PointWM's design improves the prior forecast (0.45) and gives the best Seg$_{10\%}$ (69.2). The action is actually used by the predictor ($\Delta_{\mathrm{act}}=0.10$), and its clip variance (23.8\%) is the lowest.

\begin{table}[t]
  \centering
  \footnotesize
  \setlength{\tabcolsep}{3pt}
  \fittabular{12}{lccccc}{%
    \toprule
    & \multicolumn{2}{c}{Skill at $t{+}6$} & & & \\
    \cmidrule(lr){2-3}
    Latent action & Prior & \diag{Post.} & $\Delta_{\mathrm{act}}$ & Clip var. & Seg$_{10\%}$ \\
    \midrule
    None, Table~\ref{tab:ablation}(e)        & 0.40 & --              & --        & --  & 67.0 \\
    Fixed $\mathcal{N}(\mathbf{0},\mathbf{I})$ prior & 0.40 & \diag{0.57} & 0.06 & 34.5 & 67.8 \\
    IDM on $[\mathbf{s}_{t-1};\boldsymbol{\delta}_t]$ & 0.42 & \diag{0.69} & 0.12 & 71.4 & 66.9 \\
    \rowcolor{lightblue}
    PointWM                                  & \textbf{0.45} & \diag{0.61} & 0.10 & \textbf{23.8} & \textbf{69.2} \\
    \bottomrule
  }
  \caption{Latent-action design.
  Skill over copy-last at $t{+}6$ uses the action from the context-only prior (Prior) or the posterior on the observed future (\diag{Post.}, an oracle reference).
  $\Delta_{\mathrm{act}}$ denotes the increase in latent L1 error at $t{+}6$ when the posterior comes from another clip in the batch; larger values mean the forecast depends more on the clip's own action.
  Clip var. denotes the share (\%) of posterior-mean variance between rather than within clips; lower values indicate less scene identity.
  Seg$_{10\%}$ as in Table~\ref{tab:ablation}.}
  \label{tab:latent_action}
\end{table}

\paragraph{Context-only normalization.}
Normalizing coordinates with context-only rather than full-clip statistics (Sec.~\ref{sec:objective}) barely changes the skill of PointWM (0.52 to 0.50), so forecasting does not exploit the target frames through them.

%% file: sec/5_conclusion.tex
\section{Conclusion}
\label{sec:conclusion}

We presented PointWM, the first latent world model for self-supervised learning on point-cloud videos, which turned masked latent prediction into forecasting.
Queried only with last-seen anchors and target timestamps, its predictor forecast the latent features and anchor geometry of future frames from strictly earlier context.
A key enabler was the annotation-free latent action, whose context-only prior encouraged the posterior to encode what the past left undetermined.
PointWM consistently outperformed training from scratch and leaky masked latent prediction in offline and online action segmentation and in cross-dataset action recognition.

\paragraph{Limitations and Future Work.}
Forecasts from the learned prior remained well short of those from the oracle posterior, leaving the ambiguity of an unobserved future only partly resolved, and pretraining used HOI4D alone.
Future work will strengthen the prior and scale pretraining to larger and more diverse point-cloud videos.
As an initial exploration of latent world modeling for point-cloud videos, we expect PointWM to serve as a baseline for forecasting-based representation learning.

%% file: sec/llm_usage.tex
\section*{LLM Usage Statement}
We used large language models (LLMs) as assistive tools in two parts of this
work. First, Claude (Anthropic) was used as a coding assistant to help write
and debug the experimental code, including data processing, training, and
evaluation scripts. All research ideas, method design, and experimental
protocols were conceived by the authors. Every piece of LLM-assisted code was
reviewed, tested, and run by the authors, and all reported results were
produced and verified by the authors. Second, Codex (OpenAI) was used to
polish the language of the manuscript, \eg, improving grammar, clarity, and
readability of text written by the authors. LLMs were not used to generate
research ideas, experimental results, or scientific conclusions. The authors
take full responsibility for all content of this paper, including its claims,
equations, citations, code, and results.